\documentclass[10pt,twocolumn,letterpaper]{article}

\usepackage[pagenumbers]{cvpr} 

\usepackage{graphicx}
\usepackage{mathtools}
\usepackage{amsmath}
\usepackage{amssymb}
\usepackage{booktabs}
\usepackage[table]{xcolor}
\usepackage{multirow}

\usepackage[pagebackref,breaklinks,colorlinks]{hyperref}

\usepackage[capitalize]{cleveref}
\crefname{section}{Sec.}{Secs.}
\Crefname{section}{Section}{Sections}
\Crefname{table}{Table}{Tables}
\crefname{table}{Tab.}{Tabs.}

\def\cvprPaperID{2333}
\def\confName{CVPR}
\def\confYear{2022}

\begin{document}

\title{Accelerating DETR Convergence via Semantic-Aligned Matching}

\author{
Gongjie Zhang$^1$ \qquad Zhipeng Luo$^{1,2}$ \qquad Yingchen Yu$^1$ \qquad Kaiwen Cui$^1$ \qquad Shijian Lu\thanks{\,Corresponding author.}\,\,$^{1}$ \smallskip\\
{$^{1}$Nanyang\,Technological\,University,\,Singapore \qquad $^{2}$SenseTime\,Research} \\
{\tt\small \{gongjiezhang,\,shijian.lu\}@ntu.edu.sg \qquad \{zhipeng001,\,yingchen001,\,kaiwen001\}@e.ntu.edu.sg}
}

\maketitle

\begin{abstract}

The recently developed DEtection TRansformer (DETR) establishes a new object detection paradigm by eliminating a series of hand-crafted components. However, DETR suffers from extremely slow convergence, which increases the training cost significantly. We observe that the slow convergence is largely attributed to the complication in matching object queries with target features in different feature embedding spaces. This paper presents SAM-DETR, a Semantic-Aligned-Matching DETR that greatly accelerates DETR's convergence without sacrificing its accuracy. SAM-DETR addresses the convergence issue from two perspectives. First, it projects object queries into the same embedding space as encoded image features, where the matching can be accomplished efficiently with aligned semantics. Second, it explicitly searches salient points with the most discriminative features for semantic-aligned matching, which further speeds up the convergence and boosts detection accuracy as well. Being like a plug and play, SAM-DETR complements existing convergence solutions well yet only introduces slight computational overhead. Extensive experiments show that the proposed SAM-DETR achieves superior convergence as well as competitive detection accuracy. The implementation codes are publicly available at \href{https://github.com/ZhangGongjie/SAM-DETR}{https://github.com/ZhangGongjie/SAM-DETR}\,.

\end{abstract}

\section{Introduction}   \label{sec:intro}

Object detection is one of the most fundamental tasks in computer vision and has achieved unprecedented progress with the development of deep learning~\cite{Liu2019DeepLF}. However, most object detectors often suffer from complex detection pipelines and sub-optimal performance due to their over-reliance on hand-crafted components such as anchors, rule-based target assignment, and non-maximum suppression (NMS). The recently proposed DEtection TRansformer (DETR)~\cite{DETR} removes the need for such hand-designed components and establishes a fully end-to-end framework for object detection. Despite its simple design and promising results, one of the most significant drawbacks of DETR is its extremely slow convergence on training, which requires 500 epochs to converge on the COCO benchmark~\cite{MSCOCO}, while Faster R-CNN~\cite{FasterRCNN} only takes 12$\rm\sim$36 epochs instead. This slow convergence issue significantly increases the training cost and thus hinders its more comprehensive applications.

\begin{figure}[t!] 
\begin{center}
   \includegraphics[width=0.95\linewidth]{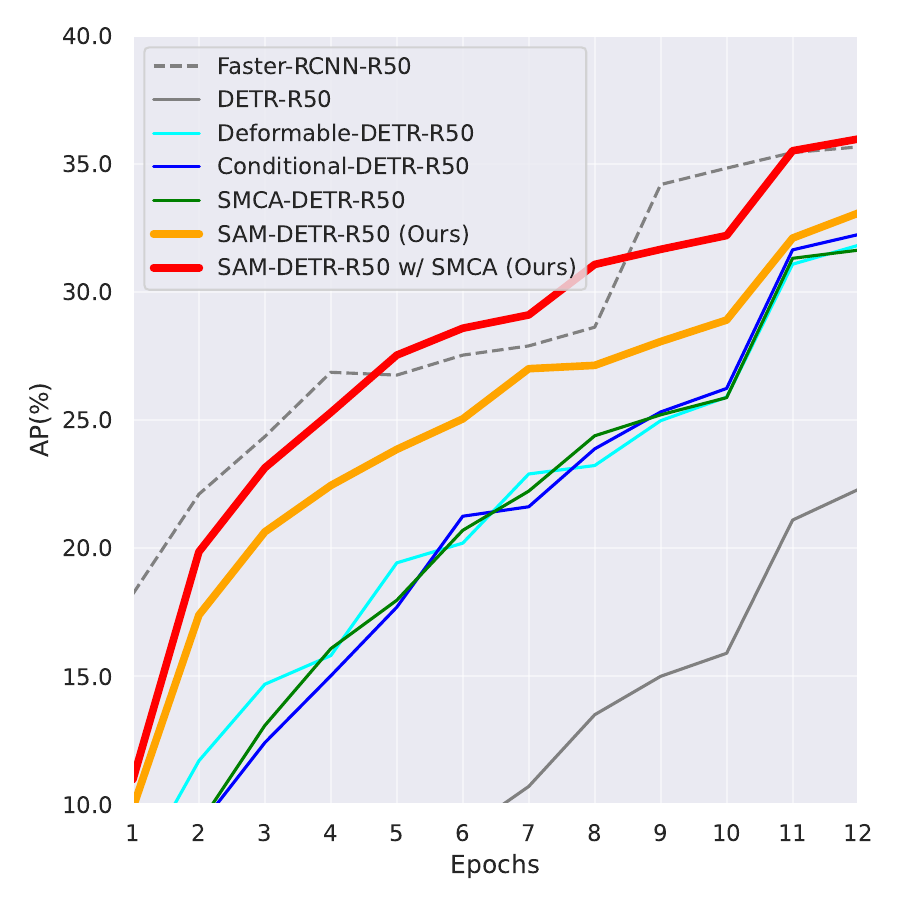}
\end{center}
\vspace*{-4.5mm}
\caption{
Convergence curves of our proposed SAM-DETR and other detectors on COCO val 2017 under the 12-epoch training scheme. All competing methods are single-scale. SAM-DETR converges much faster than the original DETR, and can work in complementary with existing convergence-boosting solutions, reaching a comparable convergence speed with Faster R-CNN.
}
\label{fig:fig1}
\vspace*{-1.0mm}
\end{figure}

\begin{figure}[t!] 
\begin{center}
   \includegraphics[width=1.0\linewidth]{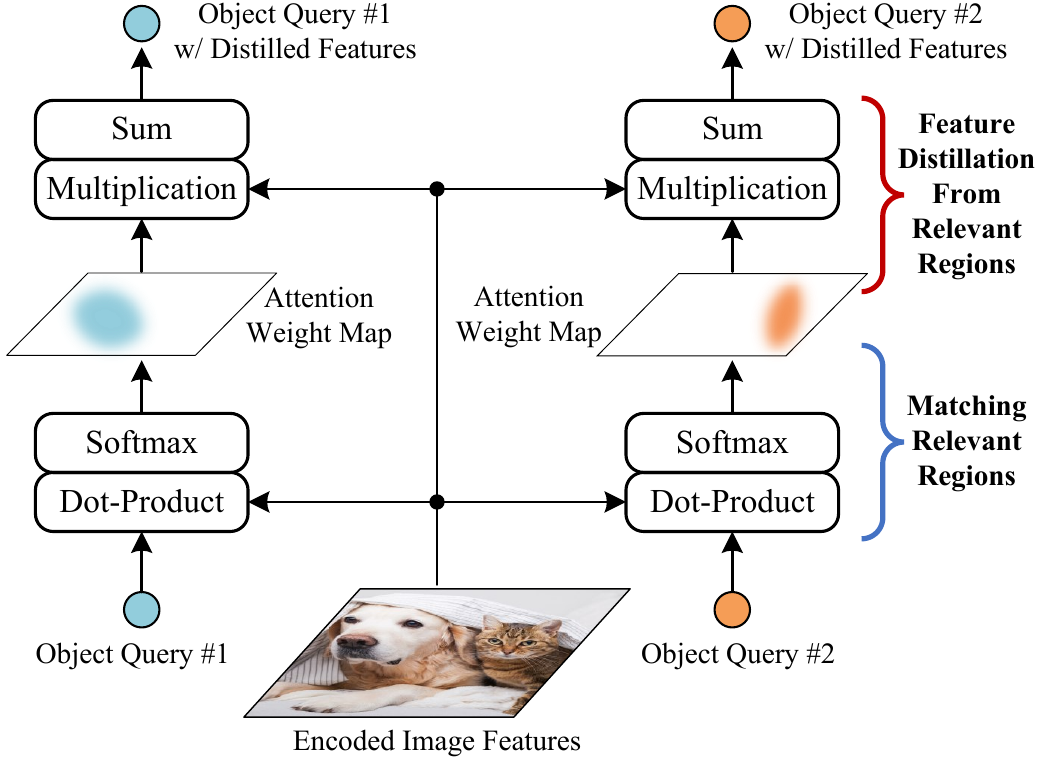}
\end{center}
\vspace*{-4.0mm}
   \caption{
   The cross-attention module in DETR's decoder can be interpreted as a `matching and feature distillation' process. Each object query first matches its own relevant regions in encoded image features, and then distills features from the matched regions, generating output for subsequent prediction.
   }
\label{fig:matching}
\vspace{-1.2mm}
\end{figure}

DETR employs a set of object queries in its decoder to detect target objects at different spatial locations. As shown in Fig.\,\ref{fig:matching}, in the cross-attention module, these object queries are trained with a set-based global loss to match the target objects and distill corresponding features from the matched regions for subsequent prediction. However, as pointed out in \cite{DeformableDETR,ConditionalDETR,SMCA-DETR}, each object query is almost equally matched to all spatial locations at initialization, thus requiring tedious training iterations to learn to focus on relevant regions. The matching difficulty between object queries and corresponding target features is the major reason for DETR's slow convergence.

A few recent works have been proposed to tackle the slow convergence issue of DETR. For example, Deformable DETR~\cite{DeformableDETR} replaces the original global dense attention with deformable attention that only attends to a small set of features to lower the complexity and speed up convergence. Conditional DETR~\cite{ConditionalDETR} and SMCA-DETR~\cite{SMCA-DETR} modify the cross-attention module to be spatially conditioned. In contrast, our approach works from a different perspective without modifying the attention mechanism.

Our core idea is to ease the matching process between object queries and their corresponding target features. One promising direction for matching has been defined by Siamese-based architecture, which aligns the semantics of both matching sides via two identical sub-networks to project them into the same embedding space. Its effectiveness has been demonstrated in various matching-involved vision tasks, such as object tracking~\cite{Siam-FC,SiamRPN,SiamRPN++,SiamRCNN,TransformerTrack,TransT}, re-identification~\cite{chung2017two,zheng2019re,wu2018and,shen2017deep,Shen_2017_ICCV}, and few-shot recognition~\cite{SiameseOneshotImageRecognition,ProtoNet,RelationNetwork,NEURIPS2019_92af93f7,MetaDETR}.
Motivated by this observation, we propose \textit{Semantic-Aligned-Matching DETR (SAM-DETR)}, which appends a plug-and-play module ahead of the cross-attention module to semantically align object queries with encoded image features, thus facilitating the subsequent matching between them. This imposes a strong prior for object queries to focus on semantically similar regions in encoded image features. In addition, motivated by the importance of objects' keypoints and extremities in recognition and localization~\cite{ExtremeNet,DETR,ConditionalDETR}, we propose to explicitly search multiple salient points and use them for semantic-aligned matching, which naturally fits in the DETR's original multi-head attention mechanism. Our approach only introduces a plug-and-play module into the original DETR while leaving most other operations unchanged. Therefore, the proposed method can be easily integrated with existing convergence solutions in a complementary manner.

\vspace{+0.88mm}
In summary, the contributions of this work are fourfold.
\textit{First}, we propose \textit{Semantic-Aligned-Matching DETR (SAM-DETR)}, which significantly accelerates DETR's convergence by innovatively interpreting its cross-attention as a `matching and distillation' process and semantically aligning object queries with encoded image features to facilitate their matching.
\textit{Second}, we propose to explicitly search for objects' salient points with the most discriminative features and feed them to the cross-attention module for semantic-aligned matching, which further boosts the detection accuracy and speeds up the convergence of our model.
\textit{Third}, experiments validate that our proposed SAM-DETR achieves significantly faster convergence compared with the original DETR.
\textit{Fourth}, as our approach only adds a plug-and-play module into the original DETR and leaves other operations mostly unchanged, the proposed SAM-DETR can be easily integrated with existing solutions that modify the attention mechanism to further improve DETR's convergence, leading to a comparable convergence speed with Faster R-CNN even within 12 training epochs.

\section{Related Work}   \label{sec:related_work}

\vspace{-0.50mm}
\noindent
\textbf{Object Detection.\;\;\;}
Modern object detection methods can be broadly classified into two categories: two-stage and single-stage detectors. Two-stage detectors mainly include Faster R-CNN~\cite{FasterRCNN} and its variants~\cite{CascadeRCNN,LibraRCNN,tychsen2018improving,RelationNetworkObjectDetection,CADNet,metarcnn,fsod,masktextspotter,FsDetView}, which employ a Region Proposal Network (RPN) to generate region proposals and then make per-region predictions over them. Single-stage detectors~\cite{SSD,YOLO9000,FewshotReweighting,RefineDet,RFBNet,zhou2019objects,incrementalfsdet,FCOS,ExtremeNet} skip the proposal generation and directly perform object classification and localization over densely placed sliding windows (anchors) or object centers. However, most of these approaches still rely on many hand-crafted components, such as anchor generation, rule-based training target assignment, and non-maximum suppression (NMS) post-processing, thus are not fully end-to-end.

\vspace{+0.2333mm}
Distinct from the detectors mentioned above, the recently proposed DETR~\cite{DETR} has established a new paradigm for object detection~\cite{DeformableDETR,DADETR,MetaDETR,PTTR,Xue2021I2C2WIT}. It employs a Transformer~\cite{transformer} encoder-decoder architecture and a set-based global loss to replace the hand-crafted components, achieving the first fully end-to-end object detector. However, DETR suffers from severe low convergence and requires extra-long training to reach good performance compared with those two-stage and single-stage detectors. Several works have been proposed to mitigate this issue: Deformable DETR~\cite{DeformableDETR} replaces the original dense attention with sparse deformable attention; Conditional DETR~\cite{ConditionalDETR} and SMCA-DETR~\cite{SMCA-DETR} propose conditioned cross-attention and Spatially Modulated Co-Attention\;(SMCA), respectively, to replace the cross-attention module in DETR's decoder, aiming to impose spatial constraints to the original cross-attention to better focus on prominent regions.  In this work, we also aim to improve DETR's convergence, but from a different perspective. Our approach does not modify the original attention mechanism in DETR, thus can work in complementary with existing methods.

\vspace{+1.25mm}
\noindent
\textbf{Siamese-based Architecture for Matching.\;\;\;}
Matching is a common concept in vision tasks, especially in contrastive tasks such as face recognition~\cite{FaceNet,song2019occlusion}, re-identification~\cite{chung2017two,zheng2019re,wu2018and,shen2017deep,Shen_2017_ICCV,li2021diverse,TransReID}, object tracking~\cite{Siam-FC,tao2016siamese,SiamRPN,SiamRPN++,SiamRCNN,TransformerTrack,TransT,dong2018triplet,he2018twofold,zhu2018distractor,zhang2019deeper,zeng2021motr}, few-shot recognition~\cite{SiameseOneshotImageRecognition,ProtoNet,RelationNetwork,NEURIPS2019_92af93f7,zhang2021pnpdet,MetaDETR}, \textit{etc}. Its core idea is to predict the similarity between two inputs. Empirical results have shown that Siamese-based architectures, which project both matching sides into the same embedding space, perform exceptionally well on the tasks involving matching. Our work is motivated by this observation to interpret DETR's cross-attention as a `matching and feature distillation' process. To achieve fast convergence, it is crucial to ensure the aligned semantics between object queries and encoded image features, \textit{i.e.}, both of them are projected into the same embedding space.

\section{Proposed Method}   \label{sec:method}

In this section, we first review the basic architecture of DETR, and then introduce the architecture of our proposed \textit{Semantic-Aligned-Matching DETR (SAM-DETR)}. We also show how to integrate our approach with existing convergence solutions to boost DETR's convergence further. Finally, we present and analyze the visualization of a few examples to illustrate the mechanism of our approach and demonstrate its effectiveness.

\subsection{A Review of DETR}

DETR~\cite{DETR} formulates the task of object detection as a set prediction problem and addresses it with a Transformer~\cite{transformer} encoder-decoder architecture. Given an image $\mathbf{I} \in \mathbb{R}^{H_{0} \times {W_{0}} \times 3}$, the backbone and the Transformer encoder produce the encoded image features $\mathbf{F} \in \mathbb{R}^{HW \times d}$, where $d$ is the feature dimension, and $H_{0}$, $W_{0}$ and $H$, $W$ denote the spatial sizes of the image and the features, respectively. Then, the encoded image features $\mathbf{F}$ and a small set of object queries $\mathbf{Q} \in \mathbb{R}^{N \times d}$ are fed into the Transformer decoder to produce detection results, where $N$ is the number of object queries, typically 100\,$\rm\sim$\,300.

In the Transformer decoder, object queries are sequentially processed by a self-attention module, a cross-attention module, and a feed-forward network (FFN) to produce the outputs, which further go through a Multi-Layer Perceptron (MLP) to generate prediction results. A good way to interpret this process is: object queries denote potential objects at different spatial locations; the self-attention module performs message passing among different object queries; and in the cross-attention module, object queries first search for the corresponding regions to match, then distill relevant features from the matched regions for the subsequent predictions. The cross-attention mechanism is formulated as:
\begin{equation}  \label{eq:1}
\mathbf{Q^\prime} = \underbrace{\overbrace{{\rm Softmax}(\frac{(\mathbf{Q W_{\rm q}})(\mathbf{F W_{\rm k}})^{\rm T}}{\sqrt{d}})}^{\text{to match relevant regions}} (\mathbf{F W_{\rm v}})}_{\text{to distill features from matched regions}},
\end{equation}
where $\mathbf{W_{\rm q}}$, $\mathbf{W_{\rm k}}$, and $\mathbf{W_{\rm v}}$ are the linear projections for query, key, and value in the attention mechanism. Ideally, the cross-attention module's output $\mathbf{Q^\prime} \in \mathbb{R}^{N \times d}$ should contain relevant information distilled from the encoded image features to predict object classes and locations.

However, as pointed out in \cite{DeformableDETR,ConditionalDETR,SMCA-DETR}, the object queries are initially equally matched to all spatial locations in the encoded image features, and it is very challenging for the object queries to learn to focus on specific regions properly. The matching difficulty is the key reason that causes the slow convergence issue of DETR.

\begin{figure*}[t!] 
\begin{center}
   \includegraphics[width=1.0\linewidth]{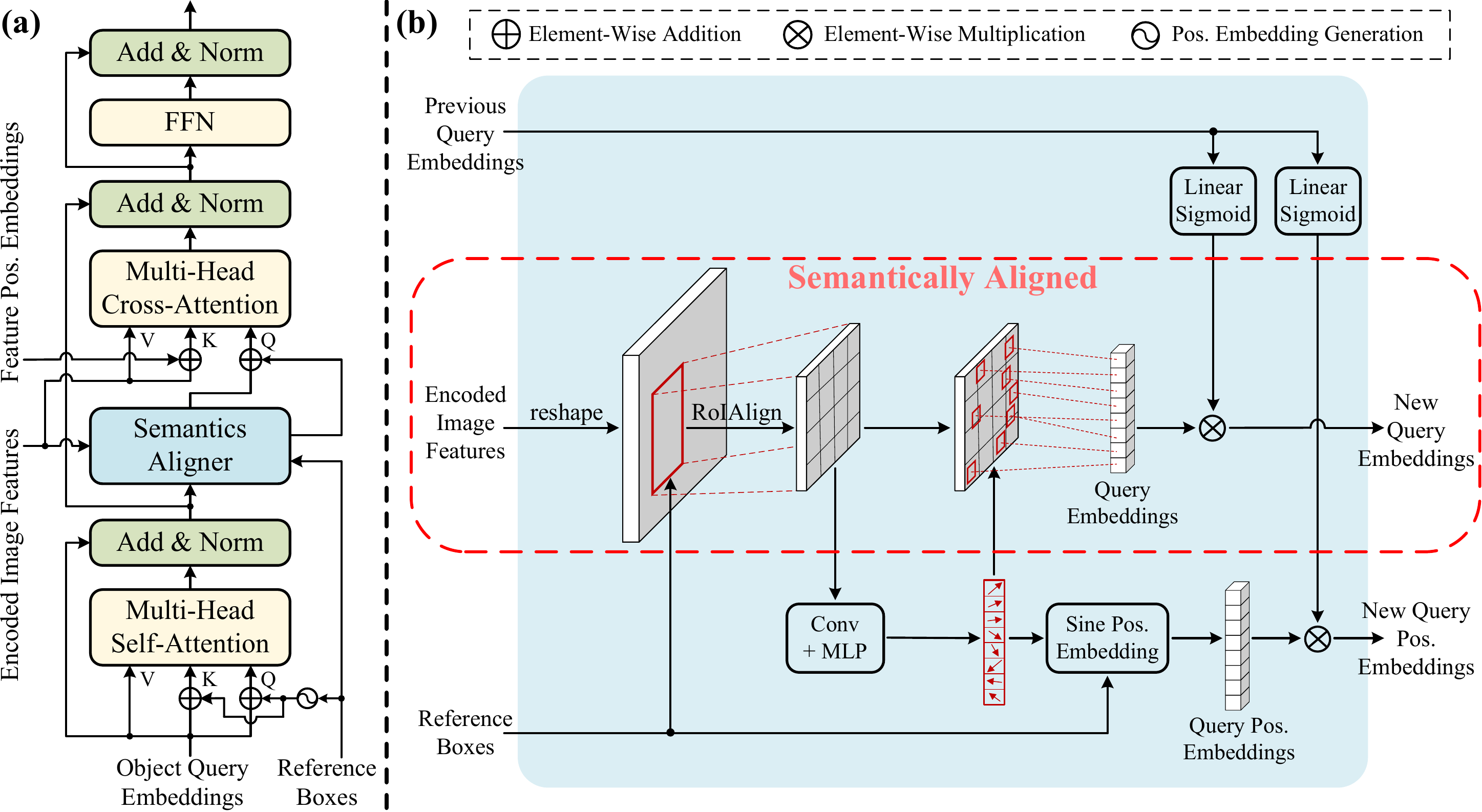}
\end{center}
\vspace*{-3.5mm}
   \caption{
   The proposed \textit{Semantic-Aligned-Matching DETR (SAM-DETR)} appends a \textit{Semantics Aligner} into the Transformer decoder layer. \textbf{(a) The architecture of one decoder layer in SAM-DETR.} It models a learnable \textit{reference box} for each object query, whose center location is used to generate corresponding position embeddings. With the guidance of the reference boxes, Semantics Aligner generates new object queries that are semantically aligned with the encoded image features, thus facilitating their subsequent matching. \textbf{(b) The pipeline of the proposed \textit{Semantics Aligner}.} For simplicity, only one object query is illustrated. It first leverages the reference box to extract features from the corresponding region via RoIAlign. The region features are then used to predict the coordinates of salient points with the most discriminative features. The salient points' features are then extracted as the new query embeddings with aligned semantics, which are further reweighted by previous query embeddings to incorporate useful information from them.
   }
\label{fig:architecture}
\vspace*{-0.0mm}
\end{figure*}

\subsection{SAM-DETR}  

Our proposed SAM-DETR aims to relieve the difficulty of the matching process in Eq.\,\ref{eq:1} by semantically aligning object queries and encoded image features into the same embedding space, thus accelerating DETR's convergence. Its major difference from the original DETR~\cite{DETR} lies in the Transformer decoder layers. As illustrated in Fig.\,\ref{fig:architecture}\,(a), the proposed SAM-DETR appends a \textit{Semantics Aligner} module ahead of the cross-attention module and models learnable \textit{reference boxes} to facilitate the matching process. Same as DETR, the decoder layer is repeated six times, with zeros as input for the first layer and previous layer's outputs as input for subsequent layers.

The learnable reference boxes $\mathbf{R}_{\rm box} \in \mathbb{R}^{N \times 4}$ are modeled at the first decoder layer, representing the initial locations of the corresponding object queries. With the localization guidance of these reference boxes, the proposed Semantics Aligner takes the previous object query embeddings $\mathbf{Q}$ and the encoded image features $\mathbf{F}$ as inputs to generate new object query embeddings $\mathbf{Q^{\rm new}}$ and their position embeddings $\mathbf{Q_{\rm pos}^{\rm new}}$, feeding to the subsequent cross-attention module. The generated embeddings $\mathbf{Q^{\rm new}}$ are enforced to lie in the same embedding space with the encoded image features $\mathbf{F}$, which facilitates the subsequent matching process between them, making object queries able to quickly and properly attend to relevant regions in the encoded image features.

\subsubsection{Semantic-Aligned Matching}

As shown in Eq.\,\ref{eq:1} and Fig.\,\ref{fig:matching}, the cross-attention module applies dot-product to object queries and encoded image features, producing attention weight maps indicating the matching between object queries and target regions. It is intuitive to use dot-product since it measures similarity between two vectors, encouraging object queries to have higher attention weights for more similar regions. However, the original DETR\,\cite{DETR} does not enforce object queries and encoded image features being semantically aligned, \textit{i.e.}, projected into the same embedding space. Therefore, the object query embeddings are randomly projected to an embedding space at initialization, thus are almost equally matched to the encoded image features' all spatial locations. Consequently, extremely long training is needed to learn a meaningful matching between them. 

With the above observation, the proposed Semantics Aligner designs a semantic alignment mechanism to ensure object query embeddings are in the same embedding space with encoded image features, which guarantees the dot-product between them is a meaningful measurement of similarity. This is accomplished by resampling object queries from the encoded image features based on the reference boxes, as shown in Fig.\,\ref{fig:architecture}\,(b).
Given the encoded image features $\mathbf{F}$ and object queries' reference boxes $\mathbf{R}_{\rm box}$, the Semantics Aligner first restores the spatial dimensions of the encoded image features from 1D sequences $HW \times d$ to 2D maps $H \times W \times d$. Then, it applies RoIAlign~\cite{MaskRCNN} to extract region-level features $\mathbf{F_{\rm R}} \in \mathbb{R}^{N \times 7 \times 7 \times d}$ from the encoded image features. The new object queries $\mathbf{Q^{\rm new}}$ and $\mathbf{Q^{\rm new}_{\rm pos}}$ are then obtained via resampling from $\mathbf{F_{\rm R}}$. More details are to be discussed in the ensuing subsection.
\begin{equation}
  \label{eq:2}
\mathbf{F_{\rm R}} = \text{RoIAlign}(\mathbf{F}, \mathbf{R}_{\rm box}) 
\end{equation}
\begin{equation}
  \label{eq:3}
\mathbf{Q^{\rm new}}, \mathbf{Q^{\rm new}_{\rm pos}} = \text{Resample}(\mathbf{F_{\rm R}}, \mathbf{R}_{\rm box}, \mathbf{Q}) 
\end{equation}
Since the resampling process does not involve any projection, the new object query embeddings $\mathbf{Q^{\rm new}}$ share the exact same embedding space with the encoded image features $\mathbf{F}$, yielding a strong prior for object queries to focus on semantically similar regions.

\subsubsection{Matching with Salient Point Features}   \label{sec:search_salient_points}

Multi-head attention plays an indispensable role in DETR, which allows each head to focus on different parts and thus significantly strengthens its modeling capacity. Besides, prior works~\cite{ExtremeNet,DETR,ConditionalDETR} have identified the importance of objects' most discriminative salient points in object detection. Inspired by these observations, instead of naively resampling by average-pooling or max-pooling, we propose to explicitly search for multiple salient points and employ their features for the aforementioned semantic-aligned matching. Such design naturally fits in the multi-head attention mechanism~\cite{transformer} without any modification.

Let us denote the number of attention heads as $M$, which is typically set to 8. As shown in Fig.\,\ref{fig:architecture}\,(b), after retrieving region-level features $\mathbf{F_{\rm R}}$ via RoIAlign, we apply a ConvNet followed by a multi-layer perception (MLP) to predict $M$ coordinates $\mathbf{R}_{\rm SP} \in \mathbb{R}^{N \times M \times 2}$ for each region, representing the salient points that are crucial for recognizing and localizing the objects. 
\vspace{-1.2333mm}
\begin{equation}
\mathbf{R}_{\rm SP} = \text{MLP}(\text{ConvNet}( \mathbf{F_{\rm R}} ))
\end{equation}
It is worth noting that we constrain the predicted coordinates to be within the reference boxes. This design choice has been empirically verified in Section\,\ref{sec:ablation_study}. Salient points' features are then sampled from $\mathbf{F_{\rm R}}$ via bilinear interpolation. The $M$ sampled feature vectors corresponding to the $M$ searched salient points are finally concatenated as the new object query embeddings, so that each attention head can focus on features from one salient point.
\vspace{-0.8mm}
\begin{equation}
\mathbf{Q}^{\rm new \prime} = \text{Concat}(\{\mathbf{F_{\rm R}}[...,x,y,...] \text{ for } x,y \in \mathbf{R}_{\rm SP}\})
\end{equation}
The new object queries' position embeddings are generated using sinusoidal functions with salient points' image-scale coordinates as input. Similarly, position embeddings corresponding to $M$ salient points are also concatenated to feed to the subsequent multi-head cross-attention module.
\vspace{-0.8mm}
\begin{equation}
\mathbf{Q}^{\rm new \prime}_{\rm pos} = \text{Concat}(\text{Sinusoidal}{(\mathbf{R}_{\rm box}, \mathbf{R}_{\rm SP})})
\end{equation}

\subsubsection{Reweighting by Previous Query Embeddings}

The Semantics Aligner effectively generates new object queries that are semantically aligned with encoded image features, but also brings one issue: previous query embeddings $\mathbf{Q}$ that contain valuable information for detection are not leveraged at all in the cross-attention module.
To mitigate this issue, the proposed Semantics Aligner also takes previous query emebddings $\mathbf{Q}$ as inputs to generate reweighting coefficients via a linear projection followed by a sigmoid function. Through element-wise multiplication with the reweighting coefficients, both new query embeddings and their position embeddings are reweighted to highlight important features, thus effectively leveraging useful information from previous query embeddings. This process can be formulated as:
\vspace{-0.8mm}
\begin{align}
\mathbf{Q}^{\rm new} = \mathbf{Q}^{\rm new \prime} \otimes \sigma (\mathbf{Q} \mathbf{W_{\rm RW1}})  \\
\mathbf{Q}^{\rm new}_{\rm pos} = \mathbf{Q}^{\rm new \prime}_{\rm pos} \otimes \sigma (\mathbf{Q} \mathbf{W_{\rm RW2}}),
\end{align}
where $\mathbf{W_{\rm RW1}}$ and $\mathbf{W_{\rm RW2}}$ denote linear projections, $\sigma(\cdot)$ denotes sigmoid function, and \,$\otimes$ denotes element-wise multiplication.

\subsection{Compatibility with SMCA-DETR}

As illustrated in Fig.\,\ref{fig:architecture}\,(a), our proposed SAM-DETR only adds a plug-and-play module with slight computational overhead, leaving most other operations like the attention mechanism unchanged. Therefore, our approach can easily work with existing convergence solutions in a complementary manner to facilitate DETR's convergence further. We demonstrate the excellent compatibility of our approach by integrating it with SMCA-DETR~\cite{SMCA-DETR}, a state-of-the-art method to accelerate DETR's convergence. 

SMCA-DETR~\cite{SMCA-DETR} replaces the original cross-attention with Spatially Modulated Co-Attention (SMCA), which estimates the spatial locations of object queries and applies 2D-Gaussian weight maps to constrain the attention responses. In SMCA-DETR~\cite{SMCA-DETR}, both the center locations and the scales for the 2D-Gaussian weight maps are predicted from the object query embeddings. To integrate our proposed SAM-DETR with SMCA, we make slight modifications: we adopt the coordinates of $M$ salient points predicted by Semantics Aligner as the center locations for the 2D Gaussian-like weight maps, and simultaneously predict the scales of weight maps from pooled RoI features. Experimental results demonstrate the complementary effect between our proposed approach and SMCA-DETR~\cite{SMCA-DETR}.

\subsection{Visualization and Analysis}

\begin{figure*}[t!] 
\begin{center}
   \includegraphics[width=0.988\linewidth]{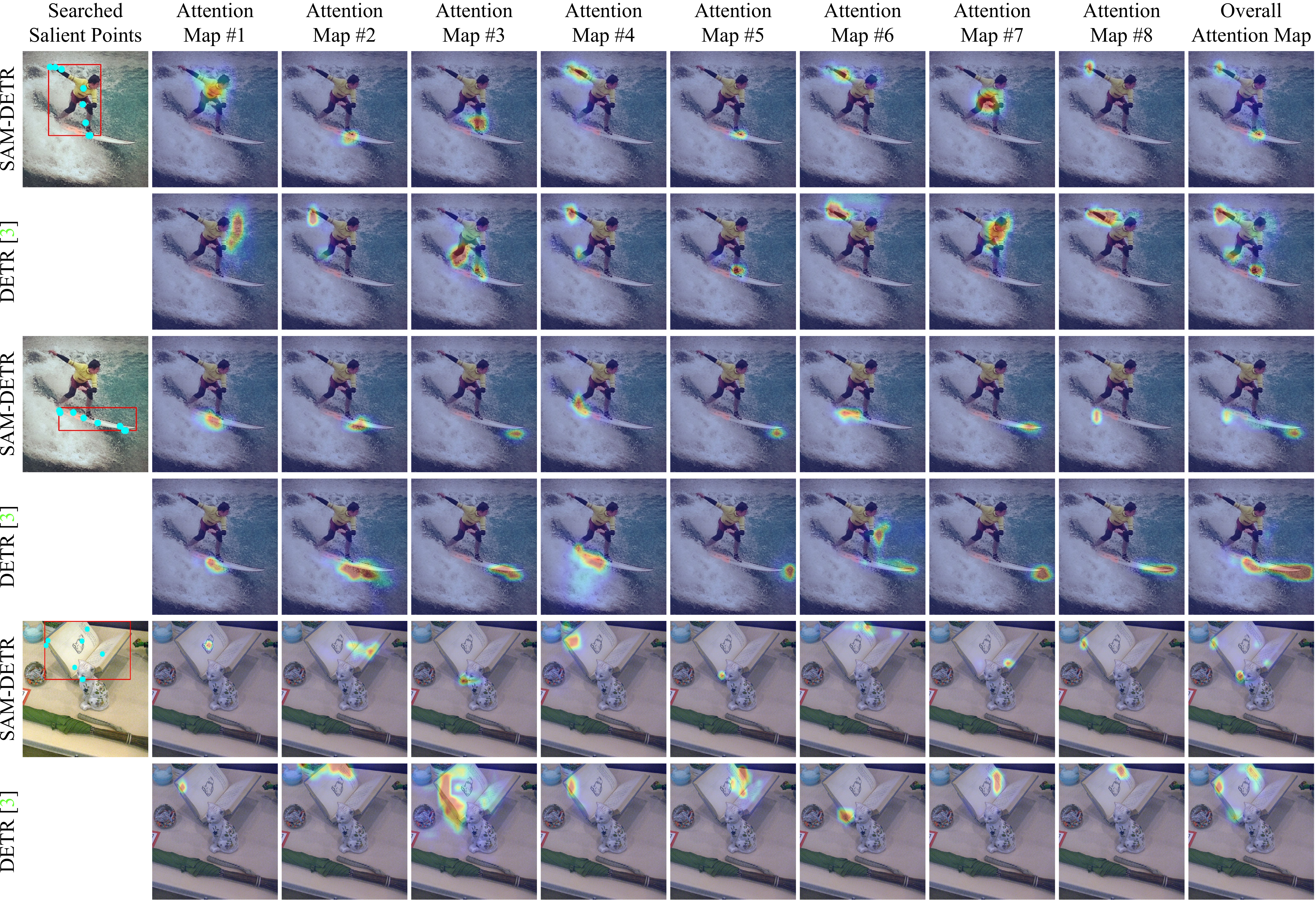}
\end{center}
\vspace*{-4.50mm}
   \caption{
   Visualization of SAM-DETR's searched salient points and their attention weight maps. The searched salient points mostly fall within the target objects and precisely indicate the locations with the most discriminative features for object recognition and localization. Compared with the original DETR, SAM-DETR's attention weight maps are more precise, demonstrating that our method effectively narrows down the search space for matching and facilitates convergence. In contrast, the original DETR's attention weight maps are more scattered, suggesting its inefficiency for matching relevant regions and distilling distinctive features.
   }
\label{fig:visualization}
\vspace*{-1.0mm}
\end{figure*}

Fig.\;\ref{fig:visualization} visualizes the salient points searched by the proposed Semantics Aligner, as well as their attention weight maps generated from the multi-head cross-attention module. We also compare them with the original DETR's attention weight maps. Both models are trained for 12 epochs with ResNet-50~\cite{resnet} as their backbones.

It can be observed that the searched salient points mostly fall within the target objects and typically are the most distinctive locations that are crucial for object recognition and localization. This illustrates the effectiveness of our approach in searching salient features for the subsequent matching process. Besides, as shown in the attention weight maps from different heads, the sampled features from each salient point can effectively match target regions and narrow down the search range as reflected by the area of attention maps. Consequently, the model can effectively and efficiently attend to the extremities of the target objects as shown in the overall attention maps, which greatly facilitates the convergence. In contrast, the attention maps generated from the original DETR are much more scattered, failing to locate the extremities efficiently and accurately. Such observation aligns with our motivation that the complication in matching object queries to target features is the primary reason for DETR's slow convergence. The visualization also proves the effectiveness of our proposed design in easing the matching difficulty via semantic-aligned matching and explicitly searched salient features.

\vspace{-1.00mm}
\section{Experiments}   \label{sec:experiments}

\begin{table*}[t]
\begin{center}
\centering
\setlength{\tabcolsep}{4.925pt}
\resizebox{1.0\textwidth}{!}{
\begin{tabular}[t]{l|c|ccc|cccccc}
\toprule[1.6666pt]
Method & multi-scale & \#Epochs & \#Params\,(M) & GFLOPs & AP & AP$_{\rm 0.5}$ & AP$_{\rm 0.75}$ & AP$_{\rm S}$ & AP$_{\rm M}$ & AP$_{\rm L}$ \\

\midrule[1.2333pt]

\multicolumn{11}{l}{\textit{Baseline methods trained for long epochs:}} \\

\midrule[0.6666pt]

Faster-RCNN-R50-DC5~\cite{FasterRCNN} & & 108 & 166 & 320 & 41.1 & 61.4 & 44.3 & 22.9 & 45.9 & 55.0 \\

Faster-RCNN-FPN-R50~\cite{FasterRCNN,FPN} & $\checkmark$ & 108 & 42 & 180 & 42.0 & 62.1 & 45.5 & 26.6 & 45.4 & 53.4 \\

DETR-R50~\cite{DETR} & & 500 & 41 & 86 & 42.0 & 62.4 & 44.2 & 20.5 & 45.8 & 61.1 \\

DETR-R50-DC5~\cite{DETR} & & 500 & 41 & 187 & 43.3 & 63.1 & 45.9 & 22.5 & 47.3 & 61.1 \\

\midrule[0.6666pt]

\multicolumn{11}{l}{\textit{Comparison of SAM-DETR with other detectors under shorter training schemes:}} \\

\midrule[0.6666pt]

Faster-RCNN-R50~\cite{FasterRCNN} & & 12 & 34 & 547 & 35.7 & 56.1 & 38.0 & 19.2 & 40.9 & 48.7 \\

DETR-R50~\cite{DETR} $\ddag$ & & 12 & 41 & 86 & 22.3 & 39.5 & 22.2 & 6.6 & 22.8 & 36.6 \\

Deformable-DETR-R50~\cite{DeformableDETR} & & 12 & 34 & 78 & 31.8 & 51.4 & 33.5 & 15.0 & 35.7 & 44.7 \\

Conditional-DETR-R50~\cite{ConditionalDETR} & & 12 & 44 & 90 & 32.2 & 52.1 & 33.4 & 13.9 & 34.5 & 48.7 \\

SMCA-DETR-R50~\cite{SMCA-DETR} & & 12 & 42 & 86 & 31.6 & 51.7 & 33.1 & 14.1 & 34.4 & 46.5 \\

\textbf{SAM-DETR-R50 (Ours)} & & 12 & 58 & 100 & 33.1 & 54.2 & 33.7 & 13.9 & 36.5 & 51.7 \\

\textbf{SAM-DETR-R50 w/\;SMCA (Ours)} & & 12 & 58 & 100 & 36.0 & 56.8 & 37.3 & 15.8 & 39.4 & 55.3 \\

\midrule[0.6666pt]

Faster-RCNN-R50-DC5~\cite{FasterRCNN} & & 12 & 166 & 320 & 37.3 & 58.8 & 39.7 & 20.1 & 41.7 & 50.0 \\

DETR-R50-DC5~\cite{DETR} $\ddag$ & & 12 & 41 & 187 & 25.9 & 44.4 & 26.0 & 7.9 & 27.1 & 41.4 \\

Deformable-DETR-R50-DC5~\cite{DeformableDETR} & & 12 & 34 & 128 & 34.9 & 54.3 & 37.6 & 19.0 & 38.9 & 47.5 \\

Conditional-DETR-R50-DC5~\cite{ConditionalDETR} &  & 12 & 44 & 195 & 35.9 & 55.8 & 38.2 & 17.8 & 38.8 & 52.0 \\

SMCA-DETR-R50-DC5~\cite{SMCA-DETR} &  & 12 & 42 & 187 & 32.5 & 52.8 & 33.9 & 14.2 & 35.4 & 48.1 \\

\textbf{SAM-DETR-R50-DC5 (Ours)} & & 12 & 58 & 210 & 38.3 & 59.1 & 40.1 & 21.0 & 41.8 & 55.2 \\

\textbf{SAM-DETR-R50-DC5 w/\;SMCA (Ours)} & & 12 & 58 & 210 & 40.6 & 61.1 & 42.8 & 21.9 & 43.9 & 58.5 \\

\midrule[0.6666pt]

Faster-RCNN-R50~\cite{FasterRCNN} & & 36 & 34 & 547 & 38.4 & 58.7 & 41.3 & 20.7 & 42.7 & 53.1 \\

DETR-R50~\cite{DETR} $\ddag$ & & 50 & 41 & 86 & 34.9 & 55.5 & 36.0 & 14.4 & 37.2 & 54.5 \\

Deformable-DETR-R50~\cite{DeformableDETR} & & 50 & 34 & 78 & 39.4 & 59.6 & 42.3 & 20.6 & 43.0 & 55.5 \\

Conditional-DETR-R50~\cite{ConditionalDETR} & & 50 & 44 & 90 & 40.9 & 61.8 & 43.3 & 20.8 & 44.6 & 59.2 \\

SMCA-DETR-R50~\cite{SMCA-DETR} & & 50 & 42 & 86 & 41.0 & - & - & 21.9 & 44.3 & 59.1 \\

\textbf{SAM-DETR-R50 (Ours)} & & 50 & 58 & 100 & 39.8 & 61.8 & 41.6 & 20.5 & 43.4 & 59.6 \\

\textbf{SAM-DETR-R50 w/\;SMCA (Ours)} & & 50 & 58 & 100 & 41.8 & 63.2 & 43.9 & 22.1 & 45.9 & 60.9 \\

\midrule[0.6666pt]

Deformable-DETR-R50~\cite{DeformableDETR} & $\checkmark$ & 50 & 40 & 173 & 43.8 & 62.6 & 47.7 & 26.4 & 47.1 & 58.0 \\

SMCA-DETR-R50~\cite{SMCA-DETR} & $\checkmark$ & 50 & 40 & 152 & 43.7 & 63.6 & 47.2 & 24.2 & 47.0 & 60.4 \\

\midrule[0.6666pt]

Faster-RCNN-R50-DC5~\cite{FasterRCNN} & & 36 & 166 & 320 & 39.0 & 60.5 & 42.3 & 21.4 & 43.5 & 52.5 \\

DETR-R50-DC5~\cite{DETR} $\ddag$ & & 50 & 41 & 187 & 36.7 & 57.6 & 38.2 & 15.4 & 39.8 & 56.3 \\

Deformable-DETR-R50-DC5~\cite{DeformableDETR} & & 50 & 34 & 128 & 41.5 & 61.8 & 44.9 & 24.1 & 45.3 & 56.0 \\

Conditional-DETR-R50-DC5~\cite{ConditionalDETR} & & 50 & 44 & 195 & 43.8 & 64.4 & 46.7 & 24.0 & 47.6 & 60.7 \\

\textbf{SAM-DETR-R50-DC5 (Ours)} & & 50 & 58 & 210 & 43.3 & 64.4 & 46.2 & 25.1 & 46.9 & 61.0 \\

\textbf{SAM-DETR-R50-DC5 w/\;SMCA (Ours)} & & 50 & 58 & 210 & 45.0 & 65.4 & 47.9 & 26.2 & 49.0 & 63.3  \\

\midrule[0.6666pt]

\multicolumn{11}{l}{\textit{Accelerating DETR's convergence with self-supervised learning:}} \\

\midrule[0.6666pt]

UP-DETR-R50~\cite{up-detr} & & 150 & 41 & 86 & 40.5 & 60.8 & 42.6 & 19.0 & 44.4 & 60.0 \\

UP-DETR-R50~\cite{up-detr} & & 300 & 41 & 86 & 42.8 & 63.0 & 45.3 & 20.8 & 47.1 & 61.7 \\

\bottomrule[1.6666pt]
\end{tabular}}
\end{center}
\vspace*{-4.888mm}
\caption{Comparison of the proposed SAM-DETR, other DETR-like detectors, and Faster R-CNN on COCO 2017 val set. $\ddag$ denotes the original DETR~\cite{DETR} with aligned setups, including increased number of object queries (100$\rightarrow$300) and focal loss for classification.}
\label{tab:exp_results}
\vspace*{-1.8mm}
\end{table*}

\vspace{-0.50mm}
\subsection{Experiment Setup}

\vspace{-0.80mm}
\noindent
\textbf{Dataset and Evaluation Metrics.\;\;\;}
We conduct experiments on the COCO 2017 dataset~\cite{MSCOCO}, which contains $\sim$117k training images and 5k validation images. Standard evaluation metrics for COCO are adopted to evaluate the performance of object detection.

\noindent
\textbf{Implementation Details.\;\;\;}
The implementation details of SAM-DETR mostly align with the original DETR~\cite{DETR}. We adopt ImageNet-pretrained\cite{imagenet} ResNet-50~\cite{resnet} as the backbone, and train our model with 8\,$\rm\times$\,Nvidia V100 GPUs using the AdamW optimizer~\cite{Adam,AdamW}.
The initial learning rate is set as $\rm 1\!\times\!10^{-5}$ for the backbone and $\rm 1\!\times\!10^{-4}$ for the Transformer encoder-decoder framework, with a weight decay of $\rm 1\!\times\!10^{-4}$. The learning rate is decayed at a later stage by 0.1. The batch size is set to 16. When using ResNet-50 with dilations (R50-DC5), the batch size is 8. Model-architecture-related hyper-parameters stay the same with DETR, except we increase the number of object queries $N$ from 100 to 300, and replace cross-entropy loss for classification with sigmoid focal loss~\cite{focalloss}. Both design changes align with the recent works to facilitate DETR's convergence~\cite{DeformableDETR,ConditionalDETR,SMCA-DETR}.

We adopt the same data augmentation scheme as DETR~\cite{DETR}, which includes horizontal flip, random crop, and random resize with the longest side at most 1333 pixels and the shortest side at least 480 pixels.

We adopt two training schemes for experiments, which include a 12-epoch scheme where the learning rate decays after 10 epochs, as well as a 50-epoch scheme where the learning rate decays after 40 epochs.

\newcolumntype{?}{!{\vrule width 1.0pt}}
\begin{table}[t]
\vspace*{-1.50mm}
\begin{center}
\centering
\setlength{\tabcolsep}{5.25pt}
\resizebox{0.478\textwidth}{!}{
\begin{tabular}[t]{c|c|c|c|c|c?ccc}
\toprule[1.2333pt]

\multirow{2}{*}{SAM} & \multicolumn{4}{c|}{\small Query Resampling Strategy} & \multirow{2}{*}{RW} & \multirow{2}{*}{AP} & \multirow{2}{*}{AP$_{\rm 0.5}$} & \multirow{2}{*}{AP$_{\rm 0.75}$} \\ 

\cline{2-5}

& \small Avg & \small Max & \small SP\,x1 & \small SP\,x8 & & & \\

\midrule[0.8pt]

& & & & & & 22.3 & 39.5 & 22.2 \\

$\checkmark$ & $\checkmark$ & & & & & 25.2 & 48.9 & 23.3 \\

$\checkmark$ & & $\checkmark$ & & & & 27.0 & 50.2 & 25.8 \\

$\checkmark$ & & & $\checkmark$ & & & 28.6 & 50.3 & 28.1 \\

$\checkmark$ & & & $\checkmark$ & & $\checkmark$ & 30.3 & 52.0 & 29.8 \\

$\checkmark$ & & & & $\checkmark$ & & 32.0 & 53.4 & 32.8 \\

$\checkmark$ & & & & $\checkmark$ & $\checkmark$ & 33.1 & 54.2 & 33.7 \\

\bottomrule[1.2333pt]

\end{tabular}
}
\end{center}
\vspace*{-5mm}
\caption{Ablation studies on our proposed design choices. Results are obtained on COCO val 2017. `SAM' denotes the proposed Semantic-Aligned Matching. `RW' denotes reweighting by previous query embeddings. Different resampling strategies for SAM are studied, including average-pooling (Avg), max-pooling (Max), one salient point (SP\,x1), and eight salient points (SP\,x8).}
\label{tab:samdetr_ablation1}
\vspace*{+0.5mm}
\end{table}

\begin{table}[t]
\begin{center}
\centering
\setlength{\tabcolsep}{6.66pt}
\resizebox{0.4186\textwidth}{!}{
\begin{tabular}[t]{c|c?ccc}
\toprule[1.2333pt]

\multicolumn{2}{c?}{\small Salient Point Search Range} & \multirow{2}{*}{AP} & \multirow{2}{*}{AP$_{\rm 0.5}$} & \multirow{2}{*}{AP$_{\rm 0.75}$} \\ 

\cline{1-2}

\small within ref box & \small within image & & & \\

\midrule[0.8pt]

$\checkmark$ & & 33.1 & 54.2 & 33.7 \\

& $\checkmark$ & 30.0 & 52.3 & 29.2 \\

\bottomrule[1.2333pt]

\end{tabular}
}
\end{center}
\vspace*{-5mm}
\caption{Ablation study on the salient point search range. Results are obtained on COCO val 2017.}
\label{tab:samdetr_ablation2}
\vspace*{-2.0mm}
\end{table}

\subsection{Experiment Results}

Table\;\ref{tab:exp_results} presents a thorough comparison of the proposed SAM-DETR, other DETR-like detectors~\cite{DETR,DeformableDETR,ConditionalDETR,SMCA-DETR,up-detr}, and Faster R-CNN~\cite{FasterRCNN}. As shown, Faster R-CNN and DETR can both achieve impressive performance when trained for long epochs. However, when trained for only 12 epochs, Faster R-CNN still achieves good performance, while DETR performs substantially worse due to its slow convergence. Several recent works~\cite{DeformableDETR,ConditionalDETR,SMCA-DETR} modify the original attention mechanism and effectively boost DETR's performance under the 12-epoch training scheme, but still have large gaps compared with the strong Faster R-CNN baseline. For standalone usage, our proposed SAM-DETR can achieve a significant performance gain compared with the original DETR baseline (+10.8\%\,AP) and outperform all DETR's variants~\cite{DeformableDETR,ConditionalDETR,SMCA-DETR}. Furthermore, the proposed SAM-DETR can be easily integrated with existing convergence-boosting methods for DETR to achieve even better performance. Combining our proposed SAM-DETR with SMCA~\cite{SMCA-DETR} brings an improvement of +2.9\%\,AP compared with the standalone SAM-DETR, and +4.4\%\,AP compared with SMCA-DETR~\cite{SMCA-DETR}, leading to performance on par with Faster R-CNN within 12 epochs. The convergence curves of the competing methods under the 12-epoch scheme are also presented in Fig.\,\ref{fig:fig1}.

We also conduct experiments with a stronger backbone R50-DC5 and with a longer 50-epoch training scheme. Under various setups, the proposed SAM-DETR consistently improves the original DETR's performance and achieves state-of-the-art accuracy when further integrated with SMCA~\cite{SMCA-DETR}.
The superior performance under various setups demonstrates the effectiveness of our approach.

\subsection{Ablation Study} \label{sec:ablation_study}
\vspace{-0.5mm}

We conduct ablation studies to validate the effectiveness of our proposed designs. Experiments are performed with ResNet-50~\cite{resnet} under the 12-epoch training scheme. 

\vspace{+1.45mm}
\noindent
\textbf{Effect of Semantic-Aligned Matching (SAM).\;\;\;}
As shown in Table\;\ref{tab:samdetr_ablation1}, the proposed SAM, together with any query resampling strategy, consistently achieves superior performance than the baseline. We highlight that even with the naive max-pooling resampling, AP$_{\rm 0.5}$ improves by 10.7\%, a considerable margin. The results strongly support our claim that SAM effectively eases the complication in matching object queries to their corresponding target features, thus accelerating DETR's convergence.

\vspace{+0.5mm}
\noindent
\textbf{Effect of Searching Salient Points.\;\;\;}
As shown in Table\,\ref{tab:samdetr_ablation1}, different query resampling strategies lead to large variance in detection accuracy. Max-pooling performs better than average-pooling, suggesting that detection relies more on key features rather than treating all features equally. This motivates us to explicitly search salient points and use their features for semantic-aligned matching. Results show that searching just one salient point and resampling its features as new object queries outperforms the naive resampling strategies. Furthermore, sampling multiple salient points can naturally work with the multi-head attention mechanism, further strengthening the representation capability of the new object queries and boosting performance.

\vspace{+0.8mm}
\noindent
\textbf{Searching within Boxes \textit{vs.} Searching within Images.\;\;\;}
As introduced in Section\;\ref{sec:search_salient_points}, salient points are searched within the corresponding reference boxes. As shown in Table\;\ref{tab:samdetr_ablation2}, searching salient points at the image scale (allowing salient points outside their reference boxes) degrades the performance. We suspect the performance drop is due to increased difficulty for matching with a larger search space. It is noteworthy that the original DETR's object queries do not have explicit search ranges, while our proposed SAM-DETR models learnable reference boxes with interpretable meanings, which effectively narrows down the search space, resulting in accelerated convergence.

\vspace{+0.8mm}
\noindent
\textbf{Effect of Reweighting by Previous Embeddings.\;\;\;}
We believe previous object queries' embeddings contain helpful information for detection that should be effectively leveraged in the matching process. To this end, we predict a set of reweighting coefficients from previous query embeddings to apply to the newly generated object queries, highlighting critical features. As shown in Table\;\ref{tab:samdetr_ablation1}, the proposed reweighting consistently boosts performance, indicating effective usage of knowledge from previous object queries.

\vspace{-0.45mm}
\subsection{Limitation}
\vspace{-1.6mm}
Compared with Faster R-CNN~\cite{FasterRCNN}, SAM-DETR inherits from DETR~\cite{DETR} superior accuracy on large objects and degraded performance on small objects. One way to improve accuracy on small objects is to leverage multi-scale features, which we will explore in the future.

\vspace{-1.4mm}
\section{Conclusion}   \label{sec:conclusion}
\vspace{-1.55mm}
This paper proposes SAM-DETR to accelerate DETR's convergence. At the core of SAM-DETR is a plug-and-play module that semantically aligns object queries and encoded image features to facilitate the matching between them. It also explicitly searches salient point features for semantic-aligned matching. The proposed SAM-DETR can be easily integrated with existing convergence solutions to boost performance further, leading to a comparable accuracy with Faster R-CNN within 12 training epochs. We hope our work paves the way for more comprehensive research and applications of DETR.

\clearpage
{\small
\bibliographystyle{ieee_fullname}
\bibliography{egbib}
}

\end{document}